\documentclass{article}

\usepackage[preprint]{preprint}   

\usepackage[utf8]{inputenc}
\usepackage[T1]{fontenc}
\usepackage{url}
\usepackage{booktabs}
\usepackage{amsfonts}
\usepackage{amsmath}
\usepackage{nicefrac}
\usepackage{microtype}
\usepackage{xcolor}
\usepackage{graphicx}
\usepackage{wrapfig}
\usepackage{placeins}
\usepackage{needspace}
\usepackage{float}
\definecolor{mydarkblue}{rgb}{0,0.08,0.45}
\usepackage[colorlinks,citecolor=mydarkblue,urlcolor=mydarkblue,linkcolor=mydarkblue]{hyperref}


\newcommand{\Dre}{\Delta_{\text{re-prefill}}}   

\title{KV-streams for Efficient Compaction \\in Agentic Reinforcement Learning
}

\author{%
  {\normalfont\sffamily\bfseries\fontsize{8}{13}\selectfont Emiliano Penaloza}$^{\diamond\,*\,1,2,5}$\textbf{,}\,\,
  {\normalfont\sffamily\bfseries\fontsize{8}{13}\selectfont Dane Malenfant}$^{\diamond\,1,3}$\textbf{,}\\
  {\normalfont\sffamily\bfseries\fontsize{8}{13}\selectfont Dheeraj Vattikonda}$^{1,4}$\textbf{,}\,\,
  {\normalfont\sffamily\bfseries\fontsize{8}{13}\selectfont Roger Creus Castanyer}$^{1,2}$\textbf{,}\,\,
  {\normalfont\sffamily\bfseries\fontsize{8}{13}\selectfont Siddarth Venkatraman}$^{1,5}$\textbf{,}\,\,
  {\normalfont\sffamily\bfseries\fontsize{8}{13}\selectfont Abhay Puri}$^{6}$\textbf{,}\,\,
  {\normalfont\sffamily\bfseries\fontsize{8}{13}\selectfont Jonathan Light}$^{2,7}$\textbf{,}\,\,
  {\normalfont\sffamily\bfseries\fontsize{8}{13}\selectfont Matthew James Sargent}$^{8,9}$\textbf{,}\,\,
  {\normalfont\sffamily\bfseries\fontsize{8}{13}\selectfont Augustine N. Mavor-Parker}$^{9}$\textbf{,}\,\,
  {\normalfont\sffamily\bfseries\fontsize{8}{13}\selectfont Massimo Caccia}$^{12}$\textbf{,}\,\,
  {\normalfont\sffamily\bfseries\fontsize{8}{13}\selectfont Lucas Caccia}$^{12}$\textbf{,}\,\,
  {\normalfont\sffamily\bfseries\fontsize{8}{13}\selectfont Glen Berseth}$^{1,5}$\textbf{,}\,\,
  {\normalfont\sffamily\bfseries\fontsize{8}{13}\selectfont Esmeralda S. Whitammer}$^{10}$\textbf{,}\,\,
  {\normalfont\sffamily\bfseries\fontsize{8}{13}\selectfont Alessandro Sordoni}$^{1}$\textbf{,}\,\,
  {\normalfont\sffamily\bfseries\fontsize{8}{13}\selectfont Minseon Kim}$^{2}$\textbf{,}\,\,
  {\normalfont\sffamily\bfseries\fontsize{8}{13}\selectfont Marc-Alexandre C\^ot\'e}$^{2}$\textbf{,}\\
  {\normalfont\sffamily\bfseries\fontsize{8}{13}\selectfont Laurent Charlin}$^{\dagger\,1,11}$\textbf{,}\,\,
  {\normalfont\sffamily\bfseries\fontsize{8}{13}\selectfont Guillaume Lajoie}$^{\dagger\,1,5}$
  \vspace{0.4em} \\
  {\small
  $^{1}$Mila \,\,
  $^{2}$Microsoft \,\,
  $^{3}$McGill University \,\,
  $^{4}$Polytechnique Montr\'eal \,\,
  $^{5}$Universit\'e de Montr\'eal \,\,
  $^{6}$ServiceNow Inc \,\,
  $^{7}$Rensselaer Polytechnic Institute \,\,
  $^{8}$University College London, University of London \,\,
  $^{9}$Vmax \,\,
  $^{10}$Edinburgh University \,\,
  $^{11}$HEC Montr\'eal \,\,
  $^{12}$Cohere
  \vspace{0.5em}\\
  \scriptsize $^\diamond$Core contributors \ \ $^\dagger$Equal advising \ \ $^*$Corresponding author: \texttt{emilianopp550@gmail.com}
  }
}

\begin{document}
\fancyhead[C]{\small KV-streams for Efficient Compaction in Agentic Reinforcement Learning}

\maketitle
\thispagestyle{neuripsfirstpagelogo}


\begin{abstract}
Scaling the horizon of agentic LLMs is bottlenecked by the need to fit ever longer context traces in GPU memory. Context compaction has been the most popular mechanism to alleviate this issue, keeping GPU memory constant for a given trace. Unfortunately, most compaction strategies rely on prefilling the LLM context many times over, hindering training throughput. To alleviate this bottleneck and enable efficient trainable compaction, we propose {\it KV-streams}, a plug-and-play strategy compatible with \emph{any} compaction strategy that substantially increases throughput while showing no evidence of hindering performance. KV-streams enable scalable compaction by streaming the KV cache forward rather than flushing it after each compaction. We show that KV-streams enable four different compaction strategies, achieving a $1.2$ to $11.3\times$ wall-clock speedup in training over re-prefill compaction. Beyond efficiency, we find that the streamed KV cache can act as a recurrent state, carrying forward information that has long since disappeared from the context. Specifically, in a controlled setting we show that, contrary to prior work, RL alone is all that is needed for this behavior to emerge. Overall, we show KV-streams to be an efficient and lightweight plug-and-play addition to any post-training pipeline.

\vspace{0.6em}
{\small\sffamily \textbf{Code Repository:} \href{https://github.com/Emilianopp/KV-streams}{\textbf{\texttt{github.com/Emilianopp/KV-streams}}}}\\
{\small\sffamily \textbf{SGLang Fork:} \href{https://github.com/Emilianopp/sglang}{\textbf{\texttt{github.com/Emilianopp/sglang}}}}
\end{abstract}

\begin{figure}[H]   
  \centering
  \includegraphics[width=0.92\linewidth]{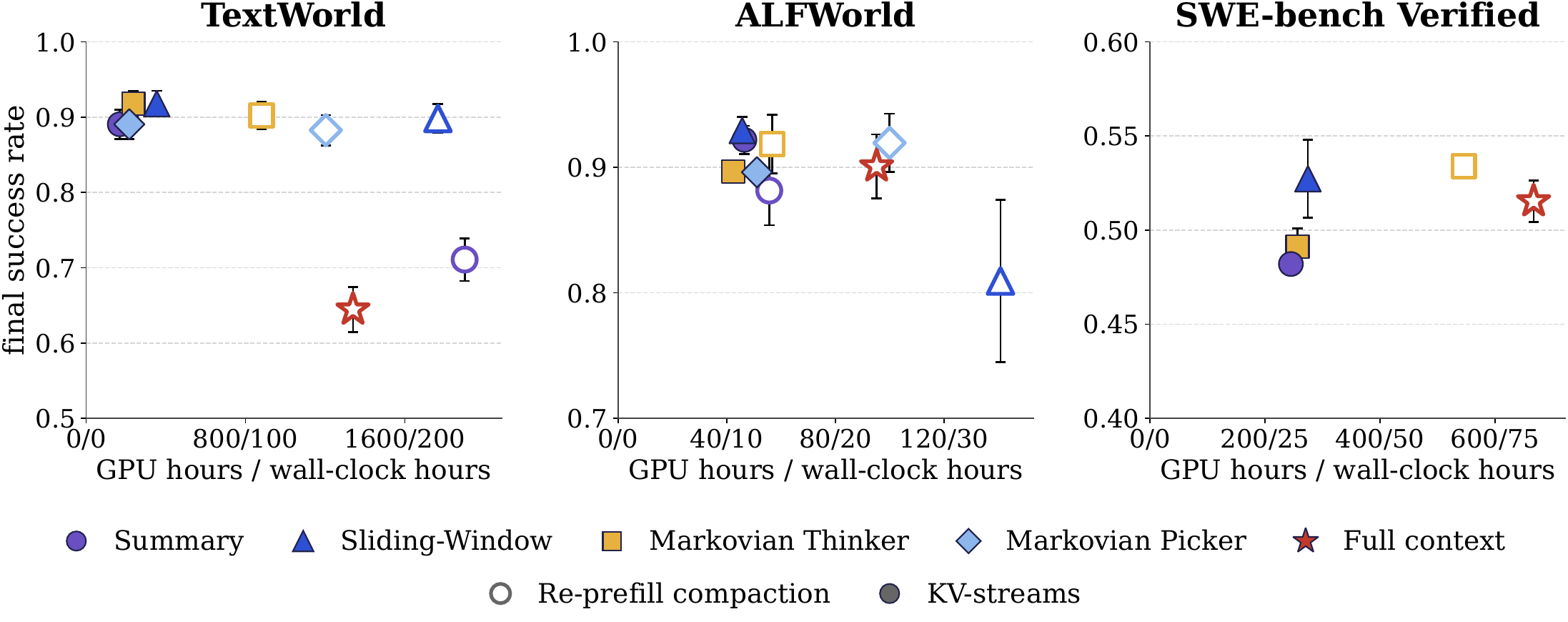}
  \caption{Final evaluation score against total training compute for every
  run on TextWorld, ALFWorld and SWE-bench Verified. Axis ticks give GPU hours
  and wall-clock hours. Marker shape and colour denote the compaction strategy. Filled markers are
  KV-streams and hollow markers re-prefill compaction, with full context (star)
  as the reference. Error bars give the standard deviation over three seeds on
  SWE-bench Verified, and the standard error over three seeds, or over evaluation
  episodes for single-seed runs, on the text-based games. KV-streams matches re-prefill compaction within error on most strategies at a
  fraction of the compute.}
  \label{fig:final-scatter}
\end{figure}

\begin{figure}[t]
  \centering
  \includegraphics[width=\linewidth]{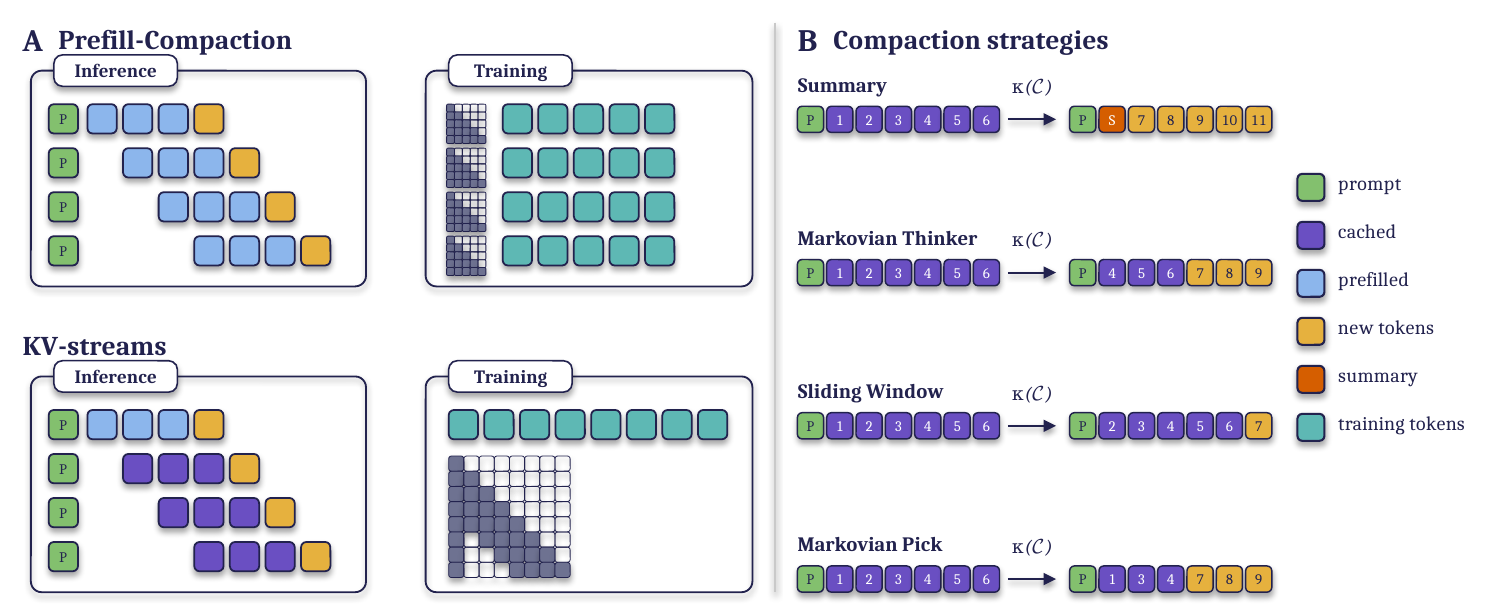}
\caption{Compaction strategies and their training costs.
\textbf{(A)} re-prefill compaction recomputes retained tokens and creates
separate training traces. KV-streams preserves retained KVs and trains
on a continuous trace with an attention mask that reproduces eviction.
\textbf{(B)} context transformations under each strategy
(P: prompt; S: summary).}
  \label{fig:schematic}
\end{figure}

\section{Introduction}
\label{sec:intro}

Reinforcement learning on long-horizon LLM agentic tasks requires generating many
concurrent rollouts, each having key-value (KV) caches growing linearly with
length. Under a fixed GPU memory budget, the longest-running rollouts quickly
become the bottleneck, monopolizing memory that shorter rollouts could
otherwise use. Reducing the memory footprint of long traces can therefore
substantially increase training throughput, enabling more optimizer steps and
larger batch sizes \citep{kwon2023efficientmemorymanagementlarge,
chen2026libraefficientresourcemanagement}. 

To process long contexts while keeping the cache size bounded, a variety of \emph{compaction} techniques can be used. They
remove or replace a portion of the context once a fixed token budget is exceeded (e.g., replacing older context with a summary)~\citep{wu2026resumunlockinglonghorizonsearch,
li2026compactionrlreinforcementlearningcontext}. Compaction lets a trace extend
indefinitely (in principle) without growing its memory footprint, but increases training compute. This is due to the retained context being prefilled each time following a  compaction, so the
model performs repeated forward passes over the same
tokens~\citep{cim2026parallelcontextcompactionlonghorizon}.

In this work, we propose {\it KV-streams} as a method to alleviate the computational bottleneck of compaction. Specifically, KV-streams evicts KV entries
directly in the inference engine, yielding the memory benefits of compaction while maintaining a single continuous stream of KVs, thus avoiding any repeated prefill cost. On the trainer side, KV-streams reproduce the
eviction pattern with a custom attention mask, so no tokens are re-processed.
Through this simple plugin, KV-streams substantially increase both training and
inference throughput on the four compaction strategies we tested and are
compatible with others. Figure~\ref{fig:schematic} summarizes those four
strategies and the mechanism behind the speedup.

We evaluate KV-streams combined with various compaction strategies on long-horizon agentic tasks, using \texttt{Qwen3-4B-Instruct-2507} and \texttt{Qwen3.5-4B},
obtaining a $3.7$ to $11.3\times$ speedup on TextWorld~\citep{côté2019textworldlearningenvironmenttextbased} and a $1.2$ to $3.1\times$ speedup on ALFWorld~\citep{shridhar2021alfworldaligningtextembodied}. Interestingly, while early in training both KV-streams and re-prefill compaction can hinder performance compared to the use of full context, with sufficient training, KV-streams can close this gap and even surpass the performance of full context. Further, we find that models trained with KV-streams also
generalize better to unseen text-based games~\citep{cui2025talestextadventurelearning,hausknecht2020interactive,scienceworld2022,NEURIPS2024_13836f25}. We then carry
these results over to agentic software-engineering tasks, where KV-streams
reach peak performance $3\times$ faster than full context while maintaining comparable
performance.

Prior work~\citep{kontonis2026mementoteachingllmsmanage} showed that preserving KVs across compactions lets the KV cache act as a ``pseudo-recurrent'' state, but found that this behaviour requires a preliminary SFT stage in their training setup. We test whether that requirement holds in our approach. Empirically, the additional wall-clock time spent on SFT does not improve performance, and synthetic experiments show that pseudo-recurrence can emerge in the KV cache without this preliminary SFT stage. This is a significant improvement for RL post-training pipelines, with up to a $5\times$ speedup.

Overall, KV-streams is a simple plug-and-play addition to any compaction strategy that greatly reduces training wall-clock, matches or exceeds the performance of re-prefill compaction, generalizes effectively, and requires no
expensive SFT stage for initialization. We make our inference and training code public at \url{https://github.com/Emilianopp/KV-streams} for a subset of the experiments.

\section{Compaction}

Long-running LLM agents commonly manage context growth by summarizing
earlier interactions or offloading information to external memory,
keeping the active context within a fixed budget
\citep{packer2024memgptllmsoperatingsystems,
wu2026resumunlockinglonghorizonsearch,
lu2025scalingllmmultiturnrl}.
It is necessary to scale LLM context windows by preventing ``context rot'' and limiting growing KV-memory. During training, limiting these effects is quite important. For instance, a single long rollout can monopolize the inference server's memory, throttling throughput. On top of that, agentic settings are specially affected by this hindrance as the environment's feedback adds to the growing context. For instance, in software engineering, reading or writing a file can consume a substantial amount of context that may not be necessary at later stages of the task. Reducing this context is therefore important for scaling long-context agentic RL. While prior works have attempted to alleviate these issues \citep{aghajohari2025markovianthinkerarchitectureagnosticlinear,wu2026reasoningcachecontinualimprovement} they either focus on non-agentic settings, which simplify the problem, they do not study its interaction with RL, or most detrimentally, they re-prefill the context incurring a high compute cost. Our goal is to fill these gaps and eliminate any redundant compute cost.

To illustrate where this added cost comes from, we formalize compaction and show how KV-streams can be applied to any strategy that meets this definition. We note that our experiments are in agentic settings, which require \textit{turn}-wise compaction (see Appendix~\ref{app:turnwise}), but we use a generalized entry-level view to describe the process, which generalizes to tokens/turns. KV-streams is readily amenable to other long-context use cases as well.

A compaction operator $\kappa$ maps a context $\mathcal{C}$ to a shorter one
$\mathcal{C}' = \kappa(\mathcal{C})$. It fires whenever a trigger $T$ is met,
either a fixed entry budget $B$ that caps $|\mathcal{C}|$ or the model's own
decision to compact \citep{li2026selfcompactinglanguagemodelagents}. When triggered, $\kappa$ composes two operators,
$\kappa = \mathrm{gen} \circ \mathrm{del}$. The \emph{delete} operator removes a
span of at least $b \ge 1$ tokens, so the context always shrinks, and \emph{generate}
emits a bridge $s$ in its place, possibly empty, such as a summary \citep{li2026compactionrlreinforcementlearningcontext, wu2026resumunlockinglonghorizonsearch}.
\label{sec:method}

\paragraph{Prefill-compaction.}
\begin{wrapfigure}{r}{0.44\textwidth}
  \centering
  \includegraphics[width=0.44\textwidth]{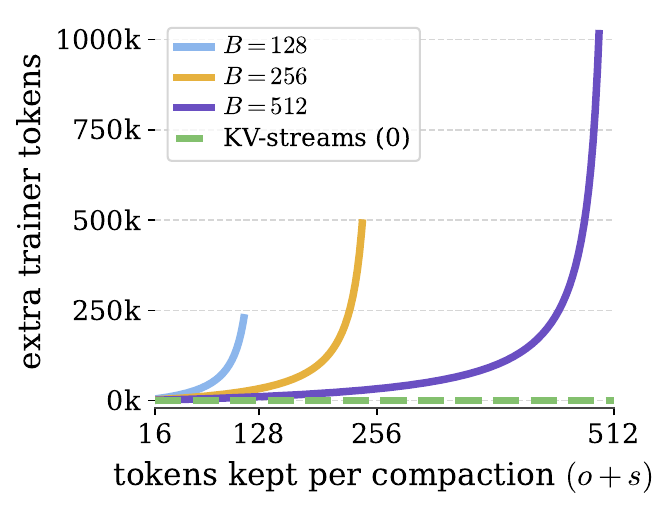}
  \caption{Extra tokens prefilled over a $32$k-token rollout vs.\ tokens kept
  per compaction $o+s$ (overlap $o$ plus summary $s$), for memory budgets $B$. As
  the retained tokens approach the budget, the repeated prefill grows by a factor of
  $\tfrac{N}{B-(o+s)}$, the number of times each kept token is re-processed by the
  trainer. KV-streams prefills nothing again (green, at $0$).}
  \label{fig:reprefill}
\end{wrapfigure}
Typical compaction pipelines continue the compacted context in a fresh trace,
which resets the kept tokens' cached KVs along with their positional
encodings. Concretely, a retained token that held position $i$ and cache
$(k_i, v_i)$ in $\mathcal{C}$ is re-emitted into the new trace $\mathcal{C}'$ with reset positional encodings. Recomputing it under this new position and context gives a different
cache entry, $(k'_i, v'_i) \neq (k_i, v_i)$, so both its position and its cached
KV change. This reset is what forces the kept entries to be prefilled
again rather than reused.

This strategy incurs a heavy cost. Let $r$ be the number of entries a compaction
keeps, the retained overlap plus any generated summary. All $r$ of them are
prefilled into the fresh trace even though their KVs were already computed
once \citep{cim2026parallelcontextcompactionlonghorizon}. Under a token/turn budget
$B$, the context holds $r$ entries right after a compaction, so the model can
generate $b = B - r$ new entries before it hits the budget and compacts again. A
rollout of $N$ entries therefore compacts $N/b$ times, and each of those
compactions prefills $r$ entries a second time. Over the whole rollout the extra
prefilled entries add up to
\[
  \Dre \;=\;
  \frac{\overbrace{N}^{\text{rollout entries}}
        \;\cdot\;
        \overbrace{r}^{\text{kept per compaction}}}
       {\underbrace{B - r}_{\text{new entries between compactions}}},
\]
which diverges as the retained context approaches the budget, derived in App.~\ref{app:reprefill}). In practice this can be mitigated by producing fewer compactions or reducing the number of entries kept per compaction (e.g. replacing all turns with a summary), yet this only mitigates the cost and does not remove it. Figure~\ref{fig:reprefill} illustrates this cost at a granular token level, showing how compacting more often makes the training grow exponentially with increased number of compactions.

\paragraph{KV-streams.}
To alleviate these issues we introduce KV-streams, a drop-in modification
that skips the repeated prefill entirely. Rather than prefilling the kept tokens
again, KV-streams carries the
retained cache forward: the retained overlap and the summary $s$ keep their
existing entries $(k_i, v_i)$ instead of being recomputed as $(k'_i, v'_i)$, so
$\Dre = 0$ . In the trainer, a custom attention mask blocks attention to the
deleted span, replicating this eviction in one forward pass. Figure~\ref{fig:schematic} visualizes the mechanisms for KV-streams, showing how by directly evicting the KV-cache a compacted rollout avoids re-prefilling and does not require to be broken into multiple traces.

\section{Compaction Strategies}
KV-streams alleviates the cost of prefilling context, but different compaction strategies retain different amounts of context $r = o + s$.
We study four that span this range, from keeping almost none of it to keeping
nearly the full budget.

\paragraph{Summary.}
The standard strategy asks the model to summarize its prior context and then
drops it \citep{wu2026resumunlockinglonghorizonsearch, li2026compactionrlreinforcementlearningcontext}. It keeps no entries ($o = 0$) and replaces them with a summary
of variable length $s$ up to a preset limit.
\paragraph{Markovian Thinker.}
The Markovian thinker~\citep{aghajohari2025markovianthinkerarchitectureagnosticlinear} keeps the prefill cost down by deleting half
the context at each compaction and replacing nothing, so $o = B/2$ and $s = 0$.
\paragraph{Markovian Pick.} It's possible that keeping the most recent half is not the most efficient strategy, rather a more efficient mechanism is to let the model pick what turns to remove and which to keep. Thus, at each compaction the model chooses $b$ entries to retain deleting all others.
\paragraph{Sliding Window.}
To probe the high-retention regime where the repeated prefill is most expensive
(Figure~\ref{fig:reprefill}), we use sliding-window compaction \citep{xiao2024efficientstreaminglanguagemodels}, which keeps almost
the full window ($o$ close to $B$, $s = 0$). It retains the most context and so
incurs the largest prefill cost, exactly where KV-streams helps most.

\section{Experimental Setup}
\label{sec:exp}

\paragraph{Benchmarks and protocol.}
We evaluate KV-streams on two text-based games and on software-engineering tasks. Both settings test whether an agent can carry information forward once older context is compacted. Text-based games often require recalling details received early in the episode, so compaction can hurt performance if the agent fails to retain them. Software-engineering tasks involve reading and writing long files, and removing these files from context can discard information the agent later relies on. The text-based games use \texttt{Qwen3-4B-Instruct-2507} and the software-engineering tasks use \texttt{Qwen3.5-4B}. Across benchmarks we sample $8$ rollouts per prompt and use a learning rate of $10^{-6}$. Within each benchmark, all runs share the same optimization settings and differ only in the compaction strategy. We report evaluation score against cumulative GPU-hours, and we stop a run early only when its wall-clock time exceeds that of the full-context run.

TextWorld \citep{côté2019textworldlearningenvironmenttextbased} provides procedurally generated games across four task families (coin\_collector, cooking, simple, treasure\_hunter) at three difficulty levels. We generate 60,000 synthetic games split uniformly across task families and difficulties, and retain 256 representative tasks for evaluation. Compaction triggers every $10$ turns, and episodes run for up to $200$ turns within a $32$k-token budget. We train for $500$ gradient steps with a batch size of $512$ on an $8$-GPU H100 node ($4$ inference, $4$ trainer), and report mean \texttt{avg@1} on the evaluation tasks.

ALFWorld \citep{shridhar2021alfworldaligningtextembodied} is a household environment in which the agent completes a language-specified task by issuing text commands and receives the admissible commands at each step. Compaction triggers every $10$ turns, and episodes run for up to $100$ turns within a $16$k-token budget. We train on the standard split for up to $200$ gradient steps with a batch size of $128$ on $4$ A100 GPUs ($1$ inference, $3$ trainer). Every $20$ gradient steps we evaluate one rollout per game on the $70$ seen and $67$ unseen validation games, reporting the mean success rate over the two splits.

For software engineering, following recent work on training small coding agents~\citep{kim2026frognanotraining4bcoding}, we train on $1{,}028$ SWE-rebench~\citep{badertdinov2025swerebench} and $443$ ScaleSWE~\citep{zhao2026scaleswe} tasks. We remove tasks that the base model solves on every attempt and tasks it never solves, measured by pass@3, so that every remaining task carries a training signal. Compaction triggers every $30$ turns, and episodes run for up to $100$ turns within a $64$k-token budget. We compare Summary, Sliding-Window and Markovian Thinker with KV-streams against Markovian Thinker with re-prefill compaction and full context. All methods train for $200$ gradient steps with a batch size of $256$ on an $8$-GPU GB200 node, and we evaluate on SWE-bench Verified~\citep{openai2024swebenchverified}.

Due to limited compute (as some experiments can take over a week to execute), we only run multi-seed experiments for AlfWorld, running single seed experiments for both software-engineering tasks and TextWorld. Appendix~\ref{app:details} lists the concurrency settings and hyperparameters for each benchmark, and Appendix~\ref{app:concurrency} describes the concurrency sweep used to choose them.

\paragraph{Inference engine implementation.}
Our aim is to show the speedup at as close to production scale as possible, so
we build on asynchronous RL frameworks that decouple the trainer and inference
GPUs. For the text-based games, we run a forked vLLM inference engine and a forked
prime-RL \citep{primeintellectteam2025intellect3technicalreport} trainer\footnote{Forked from \url{https://github.com/PrimeIntellect-ai/prime-rl}. Our code: \url{https://github.com/Emilianopp/KV-streams}.}, both modified to support KV-streams.
Our vLLM configuration uses $16$-token KV-cache blocks
\citep{kwon2023efficientmemorymanagementlarge}.
Only complete blocks are added to the prefix cache.\footnote{
\url{https://docs.vllm.ai/en/v0.9.1/design/v1/prefix_caching.html}}
When a request does not end on a multiple of $16$, the
trailing tokens stay uncommitted until a later request completes the group.
This breaks the stream, since the next request may evict tokens that the
uncommitted tokens would have attended to, and we initially found it to
substantially hinder performance. We fix it by padding each completion to a
multiple of $16$ tokens. The padding must be prefilled by both the inference
engine and the trainer, which inflates the total token count. We account for
this in Figure~\ref{fig:eval} and describe it in Appendix~\ref{app:vllm}.
Despite this inefficiency, KV-streams still improves throughput. On the
trainer side we use asynchronous RL with a maximum off-policy lag of $3$, and
we modify prime-RL to apply in-flight weight updates \emph{without} flushing
the KV cache~\citep{piché2025pipelinerlfasteronpolicyreinforcement}. This change
is important for scaling KV-streams, since flushing on every weight update
would force a prefill of entire, possibly long, traces and undo the memory
savings. For the software-engineering experiments, we instead
use a forked Slime trainer with a forked SGLang inference engine\footnote{\url{https://github.com/Emilianopp/sglang}}, which commits
individual tokens with a block size of $1$ and therefore needs no padding.

\begin{figure}[t]
  \centering
  \includegraphics[width=\linewidth]{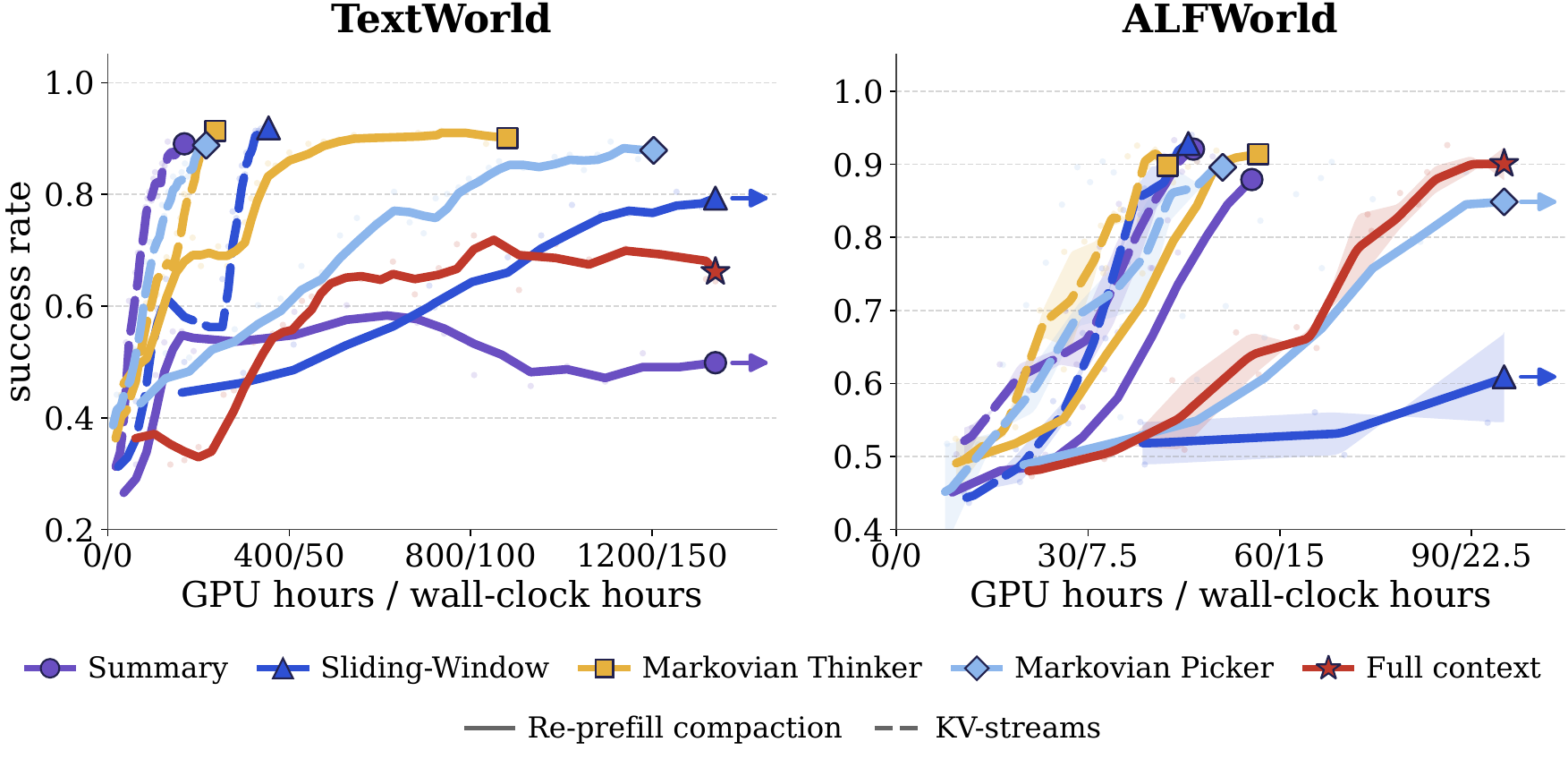}
 \caption{KV-streams reaches comparable or better success with fewer GPU hours
on TextWorld (left) and ALFWorld (right). Axis ticks give GPU hours and
wall-clock hours ($8$ GPUs on TextWorld, $4$ on ALFWorld). Solid and dashed curves denote
re-prefill compaction and KV-streams, with full context as a reference.
An arrow after the end marker shows a run truncated once it exceeds the runtime of full context. We provide complete curves
appear in Figure~\ref{fig:eval-full}.}
  \label{fig:eval}
\end{figure}

\section{Experimental Results}

    Here we describe experimental results for both text-based games and software-engineering tasks. For text-based games, we outline all compaction strategies. For software engineering, we only evaluate a subset of the compaction strategies due to compute limits.

\subsection{Text-Based Games}
\paragraph{KV-streams reduces experiment wall-time.} Figure~\ref{fig:eval} shows success rate against GPU-hours on TextWorld and
ALFWorld for every compaction strategy under re-prefill compaction and under
KV-streams. On TextWorld, KV-streams reaches the final performance of re-prefill
compaction with between $3.7\times$ and $11.3\times$ fewer GPU-hours, while performing better or similarly. Re-prefill
compaction is also expensive in absolute terms. Summary and Sliding-Window with
re-prefill compaction consume more GPU-hours than training with full context. We find that all variants of KV-streams substantially reduce wall-clock time compared to both full context and their re-prefill counterparts. On ALFWorld, due to shorter traces (the maximum sequence length is 16k), the
re-prefilling overhead is smaller, which reduces the gap. Regardless, KV-streams still achieves an increased throughput between $1.2\times$ and
$3.1\times$. KV-streams again reaches the final success rate of re-prefill compaction without incurring a performance loss across the four strategies. The full training curves for these runs, including the ones truncated in Figure~\ref{fig:eval}, are in Appendix~\ref{app:full-curves}.

\begin{figure}[t]
  \centering
  \includegraphics[width=\linewidth]{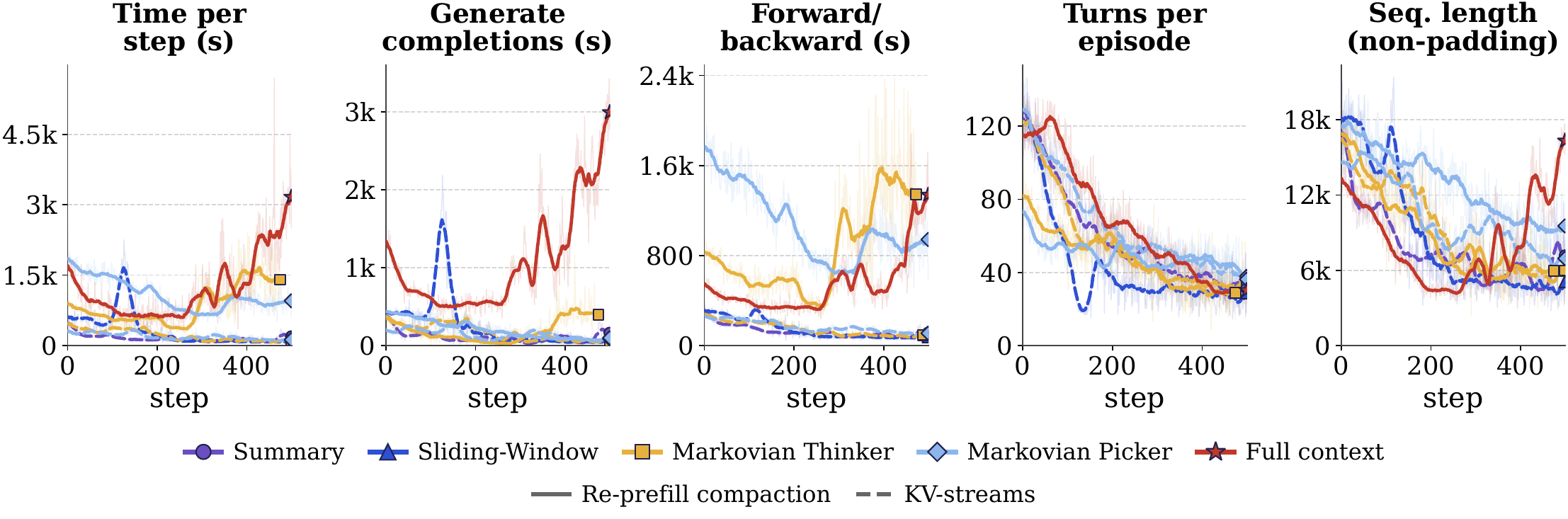}
  \caption{Smoothed per-step timing and rollout statistics on TextWorld
for runs that do not exceed the full-context wall-time. KV-streams (dashed)
reduces generation and forward/backward time relative to re-prefill
compaction (solid). Rollout statistics show mean turns per episode
and sequence length. We find KV-streams to be substantially faster during forward/backward and generation.}
  \label{fig:timing}
\end{figure}

\paragraph{Where does the speedup come from?} Figure~\ref{fig:timing} breaks the time per step into generation and forward/backward. As expected, the bottleneck for re-prefill compaction is the forward and backward pass, where a large retained context $r$ forces repeated recomputation. KV-streams instead reduces time in both the forward/backward pass and generation. While we find that initially re-prefill methods in TextWorld lead to shorter episodes and thus faster sampling, this gain diminishes as training progresses and is erased by the added training time. 

\paragraph{Performance transfer.}
We aim to evaluate whether the speedups from KV-streams come at the cost of worse transfer to other tasks. We evaluate on unseen text-based games~\citep{cui2025talestextadventurelearning,hausknecht2020interactive,scienceworld2022,NEURIPS2024_13836f25} using the final checkpoint from the \emph{TextWorld} runs, with the same budget and context settings as training ($B=10$ turns and a $32$k-token context limit). Figure~\ref{fig:transfer}A reports the score for each game averaged over three seeds. For every game, a version of KV-streams performs on par with the best strategy. A version of KV-streams also improves upon the base model in every game, showing that learning on TextWorld transfers to other text-based games. No single compaction strategy performs best across all games (Figure~\ref{fig:transfer}A), consistent with broader findings that memory-method performance depends on the task~\citep{wang2026evomembench}. Regardless, we find no loss in transfer for KV-streams compared to full context and re-prefill compaction.

\subsection{Software Engineering Agents}

\begin{figure}[t]
  \centering
  \includegraphics[width=\linewidth]{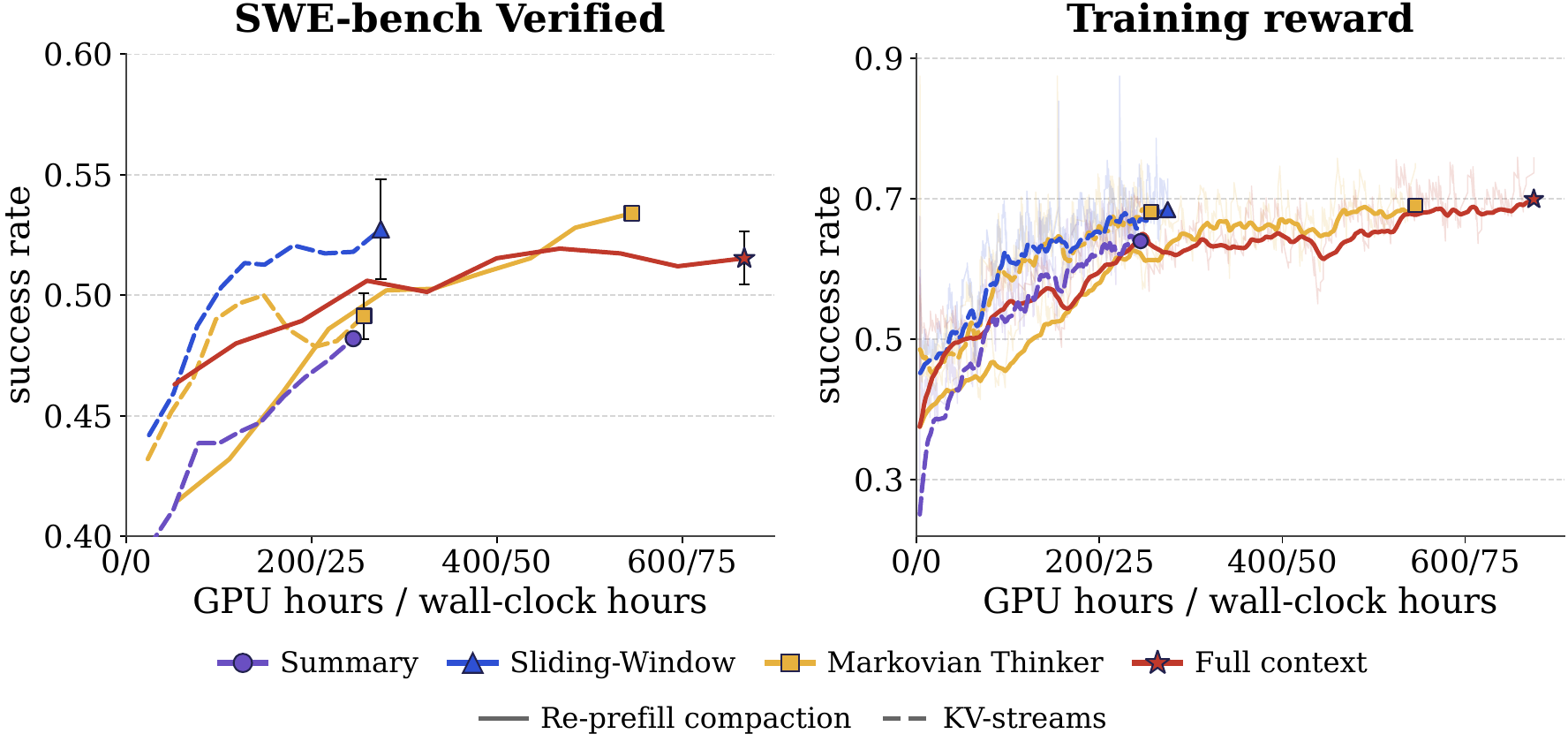}
  \caption{Comparison of KV-streams against full context and re-prefill
  Markovian Thinker on software-engineering tasks. \textbf{Left:} SWE-bench
  Verified success rate, smoothed with a centred three-checkpoint average. The
  final point is the mean and standard deviation over three evaluation seeds of
  the last checkpoint, as in Figure~\ref{fig:final-scatter}. \textbf{Right:}
  training reward, smoothed with an exponential moving average over $8$
  gradient steps for every run (raw trace faint). Axis ticks give GPU hours and
  wall-clock hours on $8$ GPUs, and every run is shown up to $200$ gradient
  steps. KV-streams reaches peak performance about $3\times$ faster than full
  context and about $2\times$ faster than re-prefill compaction.}
  \label{fig:swe}
\end{figure}

\paragraph{KV-streams scales to software-engineering tasks.} Using the training set described in Section~\ref{sec:exp}, we train for $200$ gradient steps and evaluate on SWE-bench Verified, a standard software-engineering benchmark of $500$ Python bug fixes. Figure~\ref{fig:swe} shows the evaluation results (left) and the training reward (right). To measure final performance, we evaluate the last checkpoint of each run over three seeds. Sliding-Window with KV-streams reaches $52.7 \pm 2.1\%$, statistically within the margin of error of full context ($51.5 \pm 1.2\%$) and re-prefill Markovian Thinker ($53.4 \pm 0.2\%$). Our results therefore remain consistent with the text-based games, finding no significant degradation in performance compared to full context. As before, KV-streams substantially reduces wall-clock time. Sliding-Window with KV-streams reaches its peak in about $20$ hours compared to more than $65$ hours for full context, a $3\times$ speedup. The gap persists when comparing Markovian Thinker under re-prefill compaction and under KV-streams, where KV-streams reaches its peak in about $32$ hours versus about $60$ hours for re-prefill.

\label{sec:swe}

\section{Is SFT Necessary for Recurrent KVs?}

Since retained KVs attended to tokens that are no longer in the context,
prior work has shown they can act as a psudo-recurrent state~\citep{kontonis2026mementoteachingllmsmanage}, carrying forward
information that has since been removed from the context. Because the model was
not pretrained with this in mind, prior work relies on supervised fine-tuning
to make the behaviour emerge as a first post training stage, before RL fine tuning can be done. We ask if this two-stage post training is necessary, and test whether KV-streams can enable direct RL fine tuning without an SFT stage. We test this by comparing cost savings of skipping an SFT warm start on TextWorld, and we analyze the mechanisms giving rise to this pseudo recurrent state  in a controlled synthetic experiment


\subsection{Prior SFT on TextWorld}
\label{sec:sft-textworld}

If SFT were a necessary precursor for this recurrent behaviour, we
would expect performance to be better with prior SFT. To analyze this and remove the confounder of SFT from a better model, we collect
10k trajectories from \texttt{Qwen3-4B-Instruct-2507} with a 32k context limit (matching RL training), then
perform SFT on those traces using randomly sampled eviction attention masks. For instance, sometimes we do sliding window with $B=10$ and a stride equal to 1, other times we use Markovian Thinker truncation masking half the context ($B=N/2$).

\begin{figure}[!t]
  \setlength{\abovecaptionskip}{4pt}
  \begin{minipage}[t]{0.595\linewidth}
    \raggedright\textbf{A}\par\vspace{3pt}
    \centering
    \includegraphics[width=\linewidth]{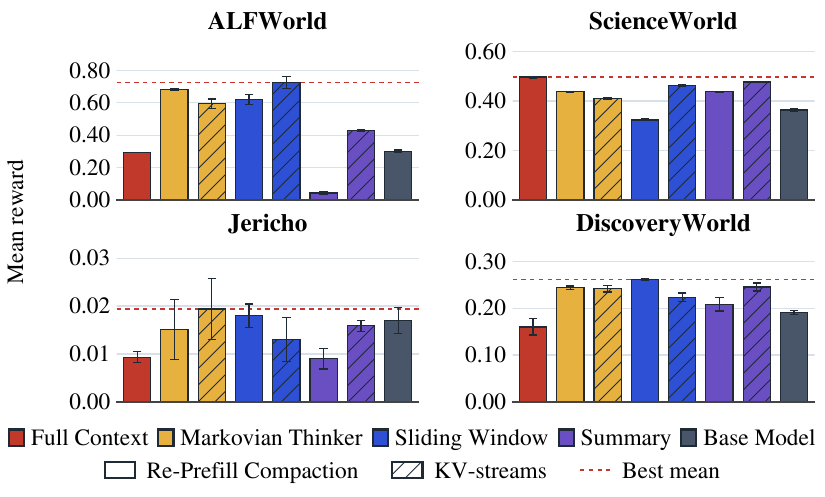}
  \end{minipage}\hfill
  \begin{minipage}[t]{0.385\linewidth}
    \raggedright\textbf{B}\par\vspace{3pt}
    \centering
    \includegraphics[width=\linewidth]{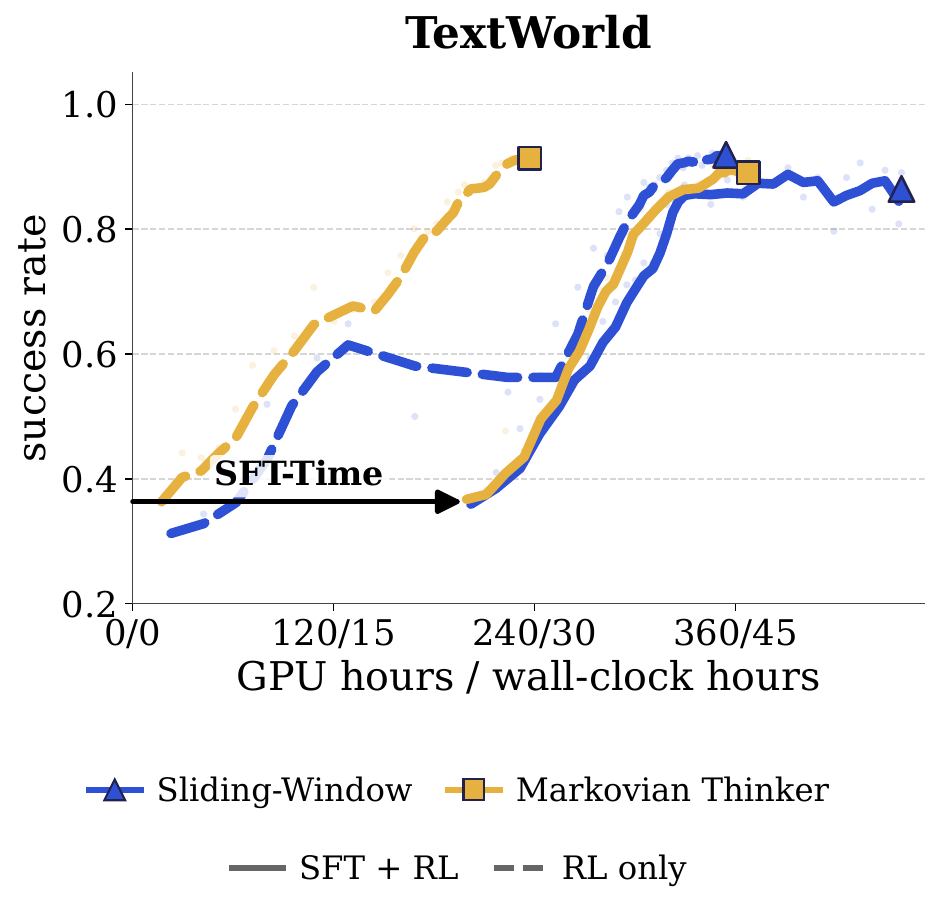}
  \end{minipage}
  \par
  \caption{(A) Transfer of TextWorld-trained checkpoints to other environments.
  Bars show mean reward and whiskers show the standard deviation across
  evaluation seeds. (B) An SFT warm-start does not pay for itself once its
  cost is counted. TextWorld success rate against GPU hours, as in
  Figure~\ref{fig:eval}, for Sliding-Window and Markovian Thinker with
  KV-streams, with an SFT warm-start followed by RL (SFT\,+\,RL, solid) and RL
  only (dashed). SFT\,+\,RL curves start after the GPU hours spent on SFT
  (SFT-Time arrow). The SFT loss is in Appendix~\ref{app:sft-loss}.}
  \label{fig:transfer}\label{fig:sft}
\end{figure}

Figure~\ref{fig:sft} B
compares RL from the base model against an SFT warm-start followed by RL~\citep{vattikonda2026trainllmwebagent}. The model fits the full-context loss (Appendix~\ref{app:sft-loss} shows the SFT training curve), yet once SFT compute is counted, RL alone is more compute efficient and shows no observable drop in performance.  We generally find that SFT does not justify its cost heavily slowing down Markovian Thinker while when using sliding window both strategies reach peak performance in the same amount of time. These results suggest that prior SFT is not an important pre-requesite for enabling KV-streams, substantially simplifying the effort needed to enable it.

\subsection{Recalling Evicted Content}
\label{sec:controlled-retrieval}
The prior experiment shows empirically that KV-streams does not depend on prior SFT. Here we test this systematically with a controlled synthetic task. The model is first assigned an object (a fruit) and then asked to decode a sequence of numbers, with a budget of $k$ tokens. Next we evict the turn containing the assignment and ask the model which object it was assigned, choosing from a list of candidates. After eviction the assignment no longer appears in the model's input, so the model cannot reference it and instructing the model to count numbers ensures it does not leak the object in its own context. Thus, it can recover the object only if the tokens it decoded while the assignment was still visible encoded that information in their KV entries, which KV-streams carries forward. Concurrent work shows that information can likewise persist in the KV cache after its source is removed from the context~\citep{dobre2026visualmemoryattackspersist}. We vary the number of decoded tokens $k\in\{16,32,64,128,256,512\}$ to test whether recall scales with the capacity of the KV-cache. Two further evaluations modify this base setting. The first asks whether a model trained to recall fruits generalizes to a new class of objects (actors). The second asks whether training can extend recall to an object the base model assigns zero probability to, by asking for a TV when the prompt lists only fruits as candidates.

\begin{figure}[!tb]
  \centering
  \setlength{\parskip}{0pt}
  \setlength{\abovecaptionskip}{4pt}
  \includegraphics[width=\linewidth]{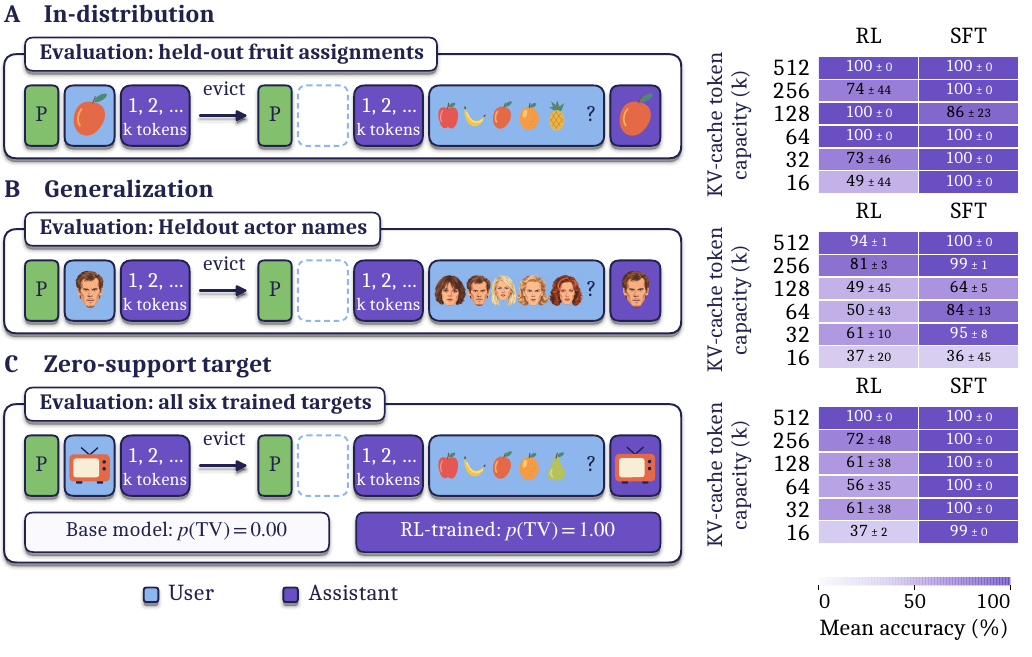}\par
\caption{Post-eviction retrieval in-distribution (A), on new names (B),
and with TV trained but unlisted (C). Heatmaps show RL and SFT accuracy
(mean $\pm$ SD over three seeds) at the best learning rate per condition. Panel C covers all
trained targets; its inset compares empirical TV-answer probabilities
in the initial and final training batches of an RL run.
Icons represent text targets.}
  \label{fig:retrieval-task}
  \label{fig:post-eviction-512}
\end{figure}

\paragraph{RL enables information leakage through KVs.} In the first task the model must retrieve the evicted assignment from a list of five candidate fruits, one of which is the assignment. Because the list always contains the answer, random sampling under RL produces some successes from the start, and the model can bootstrap from them. Figure~\ref{fig:post-eviction-512}A illustrates the setup and reports the results. As the KV-cache budget grows, recall rises to $100\%$ and matches SFT. It becomes unreliable only at the smallest budgets of $16$ and $32$ tokens. KV-streams therefore learns to carry the assignment forward through RL alone, without an SFT stage. While this result is sufficient to show that information can be leaked through KVs alone, it is restricted in the setting that it could overfit to the specific use case making things less reliable, as well as the support between the model with evicted context and unevicted context always had overlap, i.e., the both full context and evicted had sufficient probability over the assigned fruit. 

\paragraph{KV-streams generalizes across contexts.} To test whether the information carried in the KV-cache is overfit to the training context, we modify the task. We take the model trained to recall fruits and ask it to recall an actor instead. Figure~\ref{fig:post-eviction-512}B shows the results. RL and SFT perform similarly at larger token budgets, and RL trails at smaller ones. Given a sufficient KV-token budget, KV-streams trained with SFT and with RL generalizes beyond the training domain.

\paragraph{KV-streams enables recall of objects with zero initial support.} Finally, we test whether RL can teach the model to recall an object it would never sample once the assignment turn is evicted. We assign one of six objects at random, five fruits and a television, but at recall we list only the five fruits as options. The base model therefore places almost no probability on the television and would not sample it on its own. A model with full context, by contrast, can always recall that it was assigned the television. This lets us test whether information carried in the KV-cache can align the compacted model's distribution with the full-context one, bringing in a token that would otherwise never be sampled. SFT has no difficulty here by construction, since it forces the target response. Figure~\ref{fig:post-eviction-512}C shows that RL learns this behaviour as well, and as the KV-token budget grows it recalls the television $100\%$ of the time. Even in this hardest case, RL alone aligns the compacted distribution with that of full context.

Overall, our empirical and synthetic experiments show that RL alone is enough for the model to carry information forward through the KV-cache, given sufficient KV-cache capacity, and that no expensive SFT stage is needed. This broadens the applicability of KV-streams. It becomes a plug-and-play mechanism that can be added at post-training without any additional pipeline stage.  

\section{Related Work}
\label{sec:caveats}

Many methods keep long-context generation affordable by capping how much each new token can cost. Some rewrite the model itself, as in recurrent and state-space architectures, which hold cost flat but only after the model is retrained to use them~\citep{gu2024mambalineartimesequencemodeling, yang2025gateddeltanetworksimproving, fu2025slidingwindowattentiontraining}.
Transformer-XL carries hidden states between segments
\citep{dai-etal-2019-transformer}, while Compressive Transformer
compresses older memories for subsequent attention
\citep{Rae2020Compressive}. GTrXL adapts Transformer-XL for stable
reinforcement learning in memory-dependent environments
\citep{pmlr-v119-parisotto20a}.

Others leave the pretrained model untouched and simply hold the context to a fixed size, most often by summarizing old turns or keeping the most recent ones~\citep{wu2026resumunlockinglonghorizonsearch, li2026compactionrlreinforcementlearningcontext}. The second family is convenient but pays a hidden tax, since every time the budget is hit the window is torn down and rebuilt, and the same tokens are pushed through the model again~\citep{cim2026parallelcontextcompactionlonghorizon, aghajohari2025markovianthinkerarchitectureagnosticlinear, kontonis2026mementoteachingllmsmanage, wu2026reasoningcachecontinualimprovement}. That repeated work is the prefill cost we study in Section~\ref{sec:method}, and it is what makes these runs jagged and starves the GPU. KV-streams targets this tax directly. We retain whatever the strategy chooses to keep, but rather than rebuild the window we drop the evicted keys and values inside the inference engine and mirror that drop in the trainer with an attention mask. The upshot is a smooth, steady cost per token that does not depend on which compaction rule is in play, close in spirit to schemes like attention sinks~\citep{xiao2024efficientstreaminglanguagemodels} that sidestep rebuilding but without committing to one fixed retention pattern. In concurrent work, sliding windows over the KV cache have also been shown to be an efficient alternative for long generations, for test-time scaling~\citep{muennighoff2026prefixslidingefficienttesttime} and against linear attention~\citep{jolicoeurmartineau2026slidingwindowbeatslinearattention}. KV-streams is complementary, since it applies to any compaction strategy and removes the re-prefill cost of training it with RL.
Inference-time cache eviction is also studied by H$_2$O, which
retains recent and high-attention tokens \citep{3666122.3667628},
and TOVA, which interprets Transformers as multi-state RNNs and
bounds their state through KV-cache compression
\citep{oren-etal-2024-transformers}.
Our focus is on efficient agentic RL training that reproduces
the inference-time eviction pattern.

A related line of work trains the model to manage its own context, so that the content surviving a compaction is itself learned, as a generated summary in some methods and as a compact memory state in others~\citep{yan2026inftythinkbreakinglengthlimits, yan2026inftythinkeffectiveefficientinfinitehorizon, zhou2025mem1learningsynergizememory, li2026compactionrlreinforcementlearningcontext, lu2025scalingllmmultiturnrl}. These methods decide what to keep, and they all rebuild the window after each compaction, so the retained content is prefilled again and the training cost we study applies to them unchanged. KV-streams is orthogonal to this choice. It does not propose a new compaction rule and instead makes any rule cheaper to train, so these learned strategies could run on top of it.

AutoCompressors adapt pretrained language models to compress
context into learned summary vectors used as soft prompts
\citep{chevalier-etal-2023-adapting}. For multi-turn agents,
summarization-based context management can also be optimized
jointly with tool use through RL
\citep{lu2025scalingllmmultiturnrl}.

\section{Conclusion}

We introduce KV-streams, a plug-and-play mechanism that is compatible with any compaction strategy. We show that by avoiding the repeated prefill cost, KV-streams substantially reduces experiment wall-time, providing up to an $11.3\times$ speedup while often improving performance and at worst not hindering it. Further, we show that streamed KVs can act as a pseudo-recurrent state, carrying forward information that has long been removed from the context. Unlike prior work, we show that prior SFT is not a prerequisite for this effect to occur, which greatly simplifies the adoption of KV-streams. Overall, we show that KV-streams is a simple and effective way to substantially increase experiment throughput.

\subsection*{AI Use Statement}
In this work, we used generative AI tools to assist with manuscript drafting and editing, data analysis and plotting code, and the creation and refinement of figures, diagrams, and illustrative icons. We have reviewed all AI-assisted work and take responsibility for the final content, including text, claims, code, and visual artifacts.



\subsubsection*{Acknowledgments}
We thank the Mila IDT team for their support with the compute infrastructure
used in this work.
We thank Amirhossein Kazemnejad for their constructive feedback on the project.
EP acknowledges the support of the NSERC PGS-D grant and the Bourse en
intelligence artificielle provided by Universit\'e de Montr\'eal. LC
recognizes the support of NSERC, the Canada CIFAR AI Chair Program, the
Canada First Research Excellence Fund and IVADO.
GL and GB acknowledge the support of the Canada CIFAR AI Chair Program.

\bibliography{references}
\bibliographystyle{iclr2027_conference}

\FloatBarrier
\appendix
\raggedbottom
\section{Broader Impacts}
\label{sec:broader_impacts}

KV-streams lowers the cost of training long-running agents by reducing repeated
prefill. These savings may also make harmful or unauthorized automation cheaper.
Agents handling sensitive information still need access controls, monitoring,
and secure data handling. We release no new pretrained models or datasets.
KV-streams does not replace the safety policies, permissions, or privacy
protections of the underlying models and environments.

\section{Training and Implementation Details}
\label{app:details}

\subsection{Hyperparameters and Rollout Settings}
\label{app:hyperparameters}

Table~\ref{tab:hparams} lists the shared hyperparameters for each benchmark.
Text-based games use prime-RL with vLLM; software-engineering tasks use Slime
with SGLang. For each strategy, rollout concurrency is the same under re-prefill
compaction and KV-streams (Table~\ref{tab:concurrency}). Full context uses lower
concurrency because long traces occupy more memory (Appendix~\ref{app:concurrency}).
The async level limits how many policy versions a rollout can lag behind the
trainer; we drop rollouts that exceed the off-policy step limit.

\begin{table}[H]
  \centering
  \footnotesize
  \setlength{\tabcolsep}{4pt}
  \resizebox{\linewidth}{!}{%
  \begin{tabular}{lccc}
    \toprule
    & TextWorld & ALFWorld & SWE \\
    \midrule
    Model & \texttt{Qwen3-4B-Instruct-2507} & \texttt{Qwen3-4B-Instruct-2507} & \texttt{Qwen3.5-4B} \\
    GPUs (inference / trainer) & $4$ / $4$ H100 & $1$ / $3$ A100 & $4$ / $4$ GB200 \\
    Rollouts per prompt & $8$ & $8$ & $8$ \\
    Batch size (rollouts per step) & $512$ & $128$ & $256$ \\
    Gradient steps & $500$ & $200$ & $200$ \\
    Learning rate & $10^{-6}$ & $10^{-6}$ & $10^{-6}$ \\
    Weight decay & $0.01$ & $0.01$ & $0.1$ \\
    Adam $(\beta_1, \beta_2)$ & $(0.9, 0.9)$ & $(0.9, 0.9)$ & $(0.9, 0.98)$ \\
    Gradient norm clipping (max norm) & $1.0$ & $1.0$ & $1.0$ \\
    KL coefficient & $0.001$ & $0.001$ & $0.001$ \\
    Sampling temperature & $1.0$ & $1.0$ & $1.0$ \\
    Max tokens per turn & $1{,}024$ & $1{,}024$ & $8{,}192$ \\
    Sequence budget (tokens) & $32$k & $16$k & $64$k \\
    Max turns per episode & $200$ & $100$ & $100$ \\
    Compaction interval (turns) & $10$ & $10$ & $30$ \\
    KV block size (tokens) & $16$ & $16$ & $1$ \\
    Async level / off-policy steps & $1$ / $3$ & $1$ / $3$ & $1$ / $2$ \\
    Evaluation interval (steps) & $15$ & $20$ & $20$ \\
    Evaluation set & $256$ games & $70$ seen + $67$ unseen & SWE-bench Verified \\
    \bottomrule
  \end{tabular}}
  \caption{Hyperparameters shared by all runs within each benchmark.
  Compaction strategy and rollout concurrency vary between runs
  (Table~\ref{tab:concurrency}).}
  \label{tab:hparams}
\end{table}

\begin{table}[H]
  \centering
  \footnotesize
  \begin{tabular}{lccc}
    \toprule
    & TextWorld & ALFWorld & SWE \\
    \midrule
    Summary & $48$ ($192$) & $64$ ($64$) & -- \\
    Sliding-Window & $48$ ($192$) & $64$ ($64$) & $70$ ($280$) \\
    Markovian Thinker & $48$ ($192$) & $64$ ($64$) & $68$ ($272$) \\
    Markovian Picker & $48$ ($192$) & $64$ ($64$) & -- \\
    Full context & $12$ ($48$) & $16$ ($16$) & $28$ ($112$) \\
    \bottomrule
  \end{tabular}
  \caption{Concurrent rollouts per inference GPU for each strategy, with the
  total across GPUs in parentheses. Settings are the same for re-prefill
  compaction and KV-streams.}
  \label{tab:concurrency}
\end{table}

\FloatBarrier
\Needspace{0.52\textheight}
\subsection{Concurrency Sweep}
\label{app:concurrency}

\begin{figure}[H]
  \centering
  \includegraphics[width=0.75\linewidth]{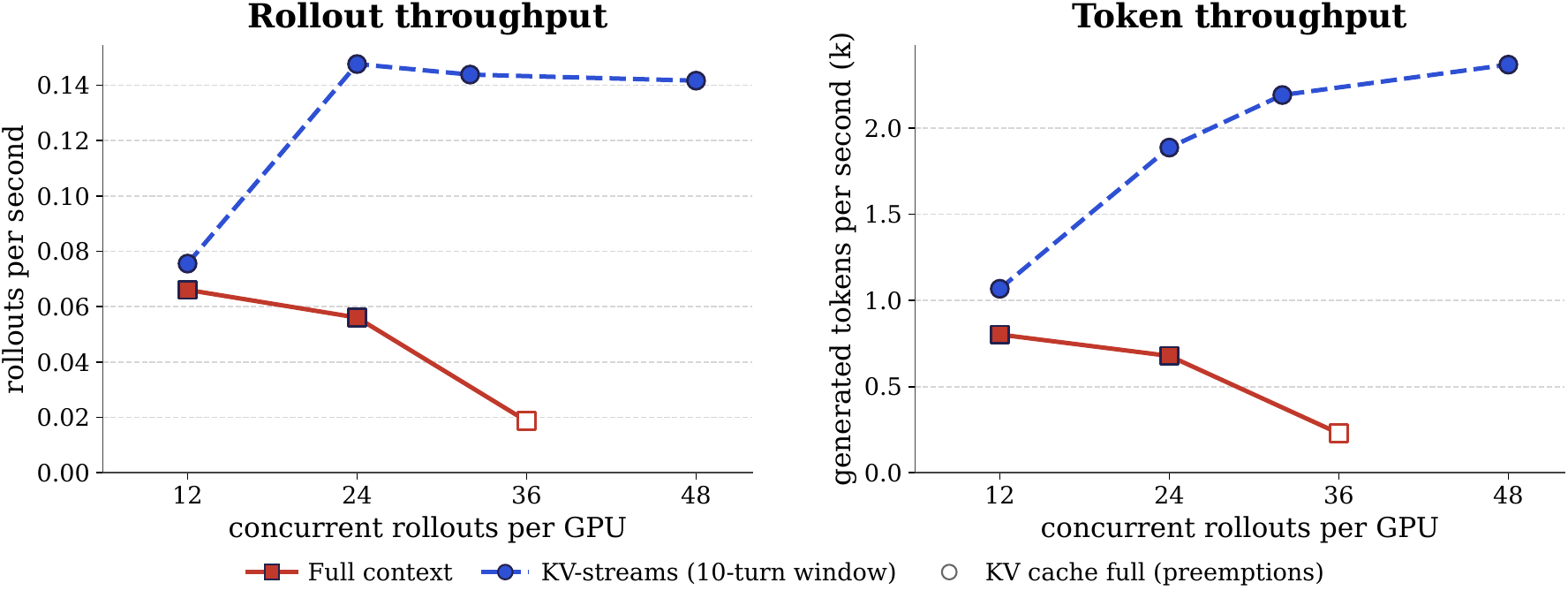}
  \caption{Sampling throughput against the number of concurrent rollouts per
  GPU on TextWorld, measured once $512$ rollouts have completed. \textbf{Left:}
  rollouts per second. \textbf{Right:} generated tokens per second. KV-streams
  uses a sliding window of $10$ turns. Hollow markers denote settings where the
  KV cache fills and vLLM preempts requests.}
  \label{fig:concurrency}
\end{figure}

We sweep rollout concurrency in vLLM on a node with four A100 (80GB) GPUs,
comparing full context with sliding-window KV-streams
(Figure~\ref{fig:concurrency}). Full-context throughput falls above $12$
concurrent rollouts per GPU, while KV-streams sustains higher concurrency.
This motivates the lower concurrency used for full context
(Table~\ref{tab:concurrency}).

\FloatBarrier
\subsection{vLLM Block Padding}
\label{app:vllm}
Our vLLM configuration commits only complete $16$-token blocks to the prefix
cache. A partial block must be recomputed on the next request. We pad each
turn to a multiple of $16$ tokens so that KV-streams can reuse all of its KVs.
The token counts in Figure~\ref{fig:eval} include this padding.
SGLang uses a block size of $1$ and needs no padding.

\subsection{Turn-Wise Compaction}
\label{app:turnwise}
In the agentic experiments, compaction removes whole turns, including their
\texttt{<|im\_start|>} and \texttt{<|im\_end|>} tags. This preserves the
boundaries of the remaining messages. The retained length $r$ counts tokens
in the retained turns and any generated summary (Section~\ref{sec:method}).

Token-level eviction produced degenerate text in our initial experiments.
The model required substantial SFT before its generations began to improve.
We suspect that removing message-boundary tokens disrupts the format learned
during pre- and mid-training. Evicting complete turns caused no such
instability in the initial responses, so we use turn-wise compaction in the
agentic experiments. It gives less precise control over context length and
per-trace memory use. Training with token-level eviction masks during
mid-training may allow finer-grained eviction.

\FloatBarrier
\newpage
\section{Additional Training Results}
\label{app:training-results}

\subsection{Full Training Curves}
\label{app:full-curves}

\begin{figure}[!hbp]
  \centering
  \includegraphics[width=0.7\linewidth]{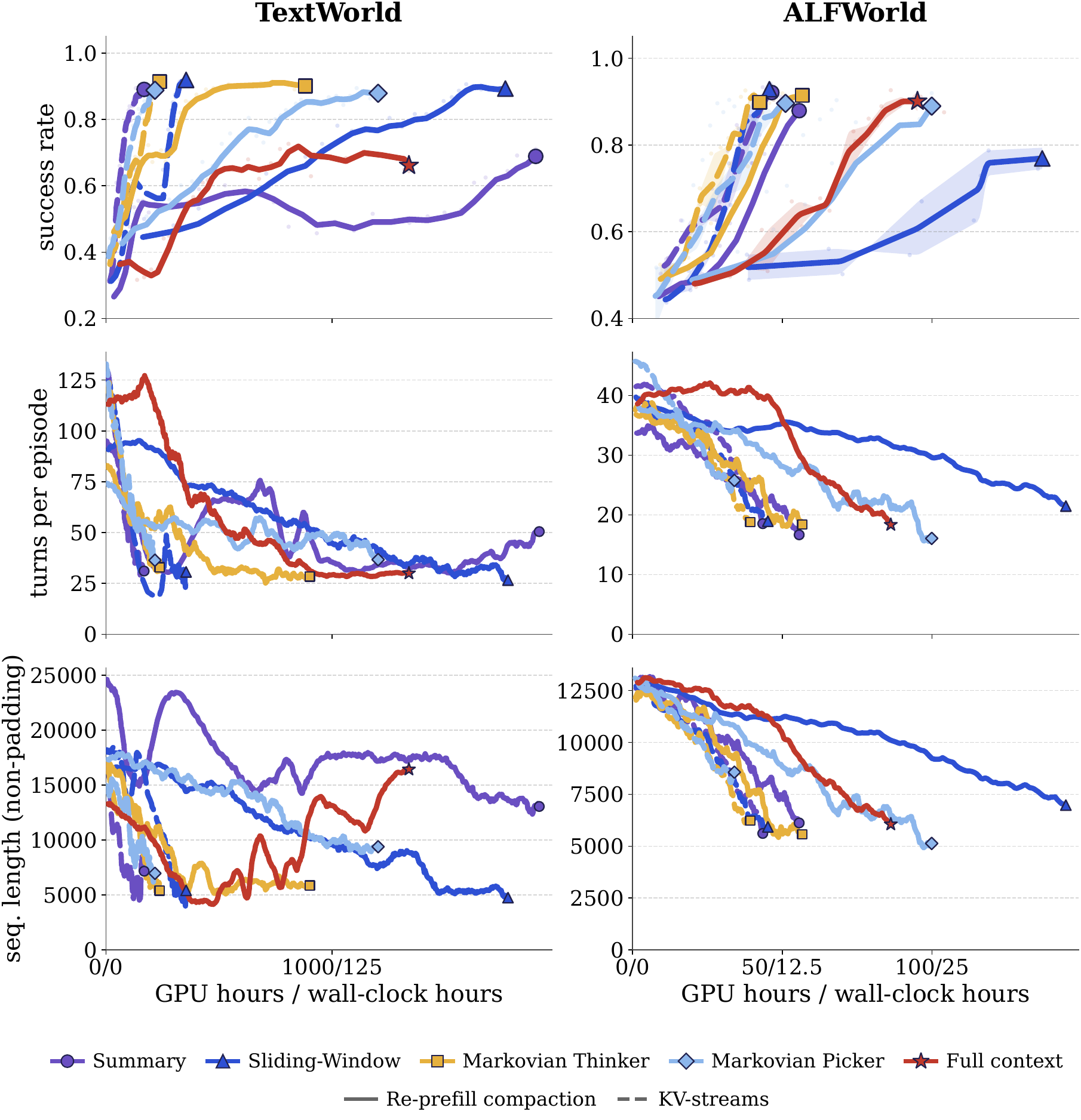}
  \caption{Training curves through each run's last checkpoint (end marker).
  Summary and Sliding-Window with re-prefill compaction exceed the full-context
  compute budget. The lower rows show smoothed mean turns per episode and
  rollout length, excluding padding.}
  \label{fig:eval-full}
\end{figure}

\begin{figure}[!hbp]
  \centering
  \includegraphics[width=\linewidth]{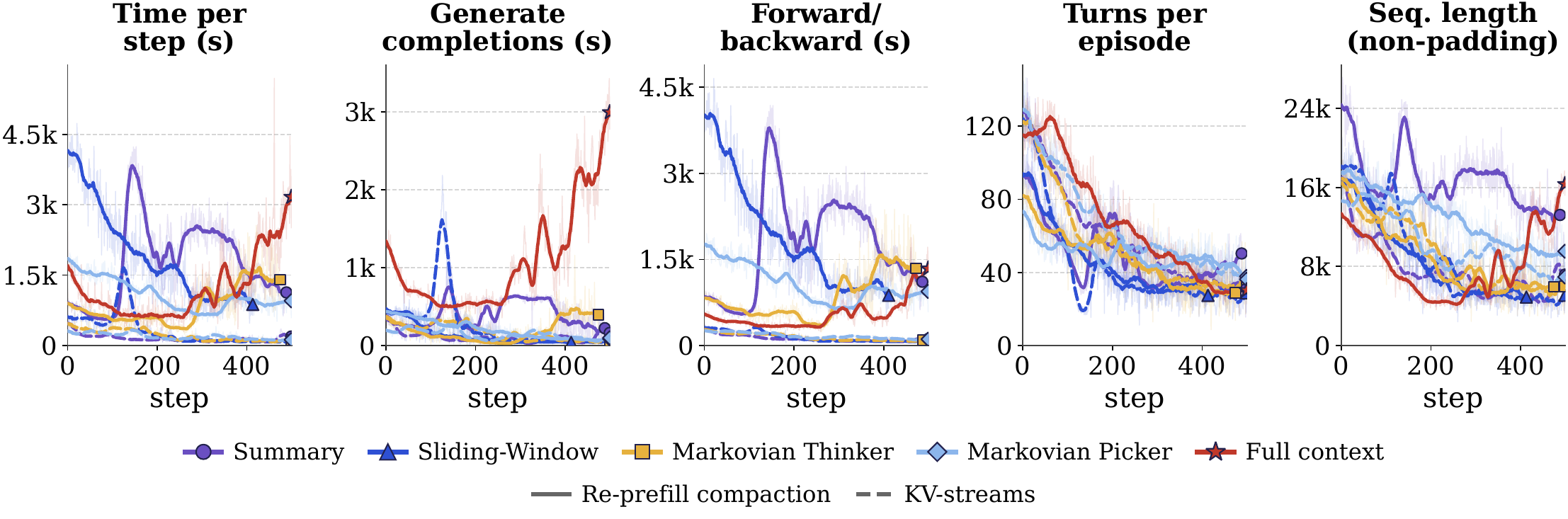}
  \caption{Per-step timing and rollout statistics, including Summary and
  Sliding-Window with re-prefill compaction.}
  \label{fig:timing-full}
\end{figure}

\begin{figure}[!hbp]
  \centering
  \includegraphics[width=\linewidth]{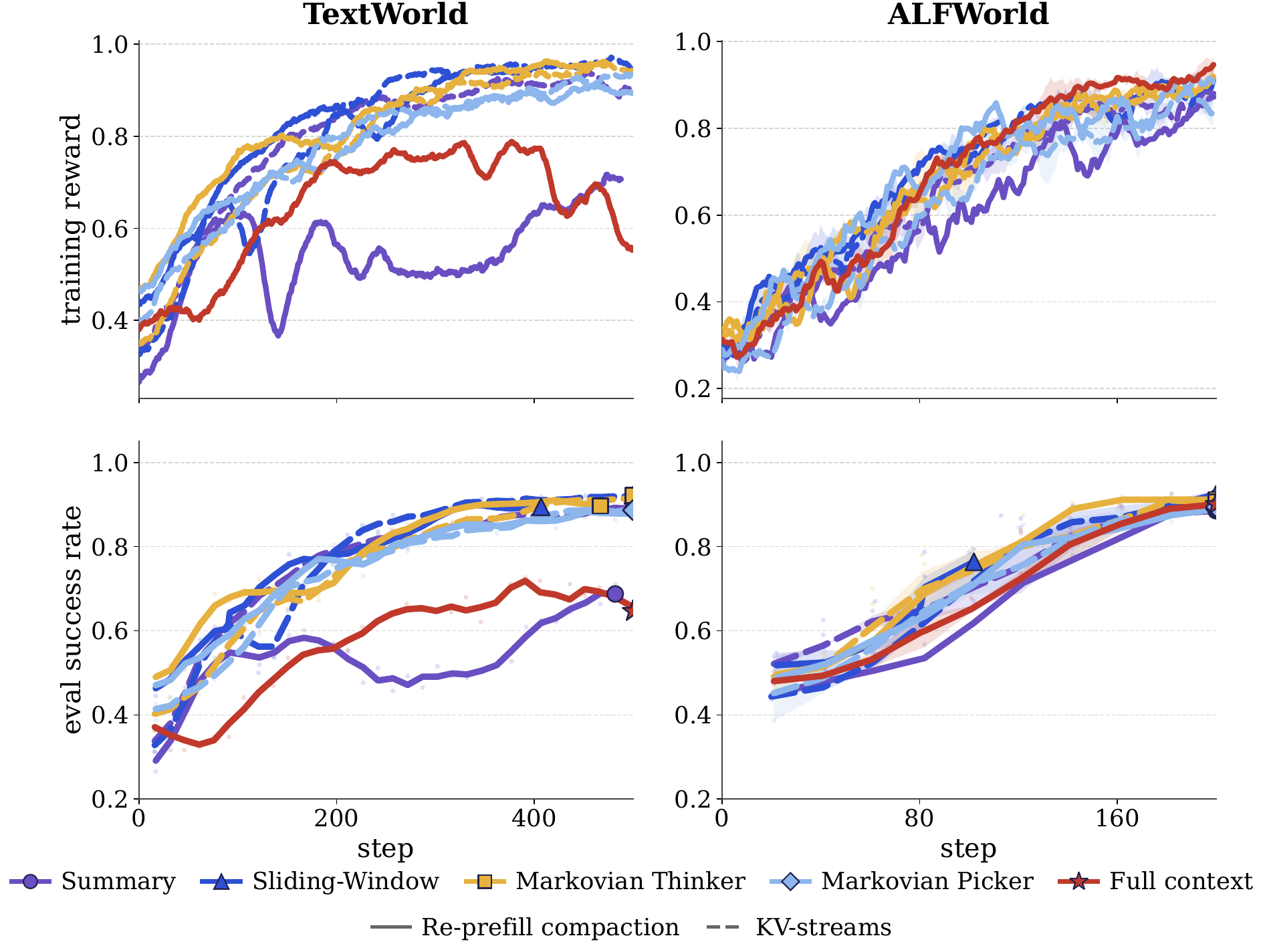}
  \caption{Training curves by gradient step on TextWorld (left) and ALFWorld
  (right). Top: smoothed mean training reward. Bottom: evaluation success rate,
  with faint points for individual evaluations. Multi-seed curves show the
  per-step mean and shaded seed range. Solid: re-prefill compaction;
  dashed: KV-streams.}
  \label{fig:reward-per-step}
\end{figure}

Figures~\ref{fig:eval-full} and~\ref{fig:timing-full} extend the training and
timing curves to each run's last checkpoint, including runs stopped after
exceeding the full-context wall-clock time. Summary and Sliding-Window with
re-prefill compaction exceed this budget and have longer step times.
Figure~\ref{fig:reward-per-step} plots mean training reward and evaluation
success rate by gradient step.

\FloatBarrier
\subsection{SFT Warm-Start Training Loss}
\label{app:sft-loss}

\begin{figure}[!hbp]
  \centering
  \includegraphics[width=0.6\linewidth]{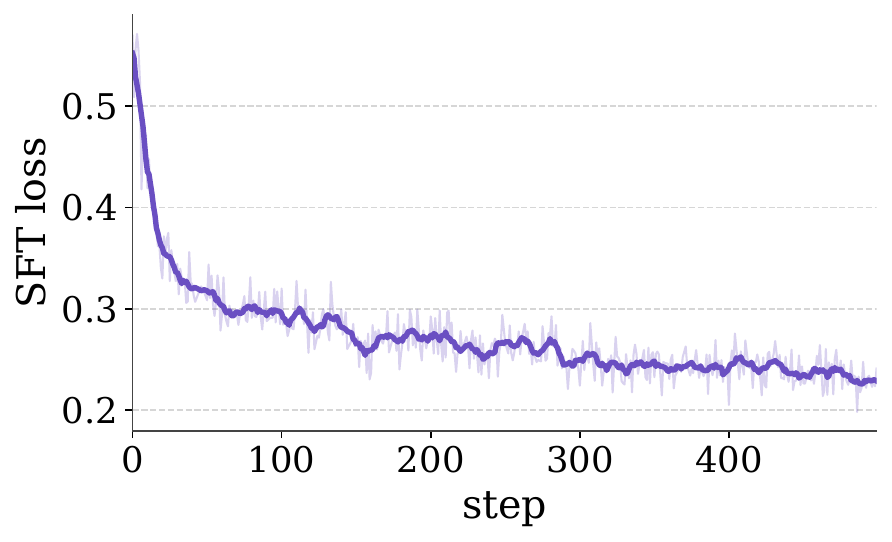}
  \caption{SFT warm-start loss by optimizer step. The faint line shows per-step
  loss; the bold line shows a centered $9$-step rolling mean.}
  \label{fig:sft-loss}
\end{figure}

The SFT warm-start in Section~\ref{sec:sft-textworld} uses $10$k full-context
trajectories from \texttt{Qwen3-4B-Instruct-2507} and randomly sampled eviction
attention masks. We train for $500$ steps. Loss falls quickly over the first
$50$ steps, then decreases more slowly (Figure~\ref{fig:sft-loss}).

\FloatBarrier
\section{Controlled Retrieval Protocol}
\label{app:fruit-retrieval}

\subsection{Task Setup}
\label{app:retrieval-setup}

\paragraph{Assignment.}
We assign the model one of five fruits: apple, banana, mango, orange,
or pineapple. The assignment prompt lists all five candidates and asks the
model to reply ``Assignment acknowledged.'' without repeating the fruit.
We evaluate on $200$ held-out trials, balanced at $40$ trials per fruit.

\paragraph{KV-cache token capacity.}
The model then decodes space-separated ascending integers, such as
``1 2 3 4 5 $\ldots$'', without naming or paraphrasing a fruit or answering
the retrieval question. We vary $k$ over $\{16,32,64,128,256,512\}$ decoded
tokens, not integers. This capacity counts only the sequence-token KVs;
prompt and other token KVs are additional. We check that the retained
sequence contains no candidate answer. All reported held-out trials pass this check.

\paragraph{Eviction and retrieval.}
We evict the assignment and acknowledgment as one turn and retain the decoded
sequence and its KVs. The model then recalls its assigned fruit from the five
candidates in the format ``Recall: \textless fruit\textgreater''.
We score the parsed answer by exact match and verify that eviction preceded
the first answer token. With no assignment or candidate answer in the retained
sequence, correct recall requires information carried forward in the KVs.

\subsection{Evaluation Conditions}
\label{app:retrieval-evaluation}

\paragraph{Prompt generalization.}
We evaluate the five-fruit checkpoints with a different retrieval prompt and
no further training (Figure~\ref{fig:post-eviction-evaluations}B). The
assignments, candidates, KV-cache token capacity, and eviction procedure are
unchanged. In both prompts, \texttt{[choices]} is a shuffled, comma-separated
list of the five fruits.
\begin{quote}
\small
\textit{Training and in-distribution evaluation:}
``Which fruit were you assigned from these five fruits: [choices]? Respond with
exactly 'Recall: \textless fruit\textgreater', replacing
\textless fruit\textgreater\ with the assigned fruit. Do not explain.''

\textit{Prompt-generalization evaluation:}
``What is your favourite object? Choose from these five options: [choices].
Respond with exactly 'Recall: \textless object\textgreater', replacing
\textless object\textgreater\ with one option. Do not explain.''
\end{quote}

\paragraph{Zero-support target (TV).}
We train on six objects: apple, banana, mango, orange, pear, and television
(Figure~\ref{fig:post-eviction-evaluations}C). The assignment and retrieval
prompts list only the five fruits, but TV assignments explicitly name
television. Sequence generation and eviction are unchanged. Correct recall
must recover the assigned object even when it is not listed.
Evaluation uses $240$ trials, balanced at $40$ per object.

\paragraph{Generalization to actor names.}
We evaluate the five-fruit checkpoints on actor names without further training:
Kathryn Erbe, Kevin Bacon, Naomi Watts, Nicole Kidman, and Toni Collette.
The model decodes the sequence, then recalls the name after we evict the
assignment turn. We evaluate with and without the five names listed as candidates,
using $200$ trials per condition, balanced at $40$ per name. As in the fruit
task, we reject sequences containing a candidate answer and score recall by
exact match.

\subsection{Training Objectives}
\label{app:retrieval-training}

\paragraph{Training data and checkpoint selection.}
We train one model per objective and KV-cache token capacity
$k\in\{16,32,64,128,256,512\}$. The five-fruit task has $1{,}000$ training
and $200$ held-out assignments; the six-target TV task has $1{,}200$ and $240$.
Both use $200$ training and $40$ held-out examples per target.
RL, RFT, and SFT start from \texttt{Qwen3-4B-Instruct} and run for $100$
updates. Self-distillation starts from the corresponding SFT checkpoint and
runs for another $100$ updates. We evaluate final checkpoints
(Figure~\ref{fig:post-eviction-evaluations}).

\paragraph{RL.}
We sample on-policy batches of $128$ trajectories, with four rollouts per
assignment, at temperature $1.0$ and top-$p=0.95$. Reward is one for an exact
post-eviction match with no candidate answer in the decoded sequence, and
zero otherwise. The loss weights each generated response token's negative
log-probability by its trajectory's advantage. The advantage uses a
batch-wide leave-one-out baseline: the mean reward of the other $127$
trajectories. If every answer in the batch is incorrect, the baseline is
chance accuracy ($1/5$ for five fruits and $1/6$ for the six-target TV task).
We add no supervised target loss.

\paragraph{RFT.}
Rejection-sampling fine-tuning (RFT) uses the same sampling and reward as RL.
It minimizes negative log-likelihood on generated response tokens from
successful trajectories. Incorrect trajectories and those containing a
candidate answer in the decoded sequence contribute no loss. RFT uses no
leave-one-out advantage weighting.
Both RL and RFT use AdamW with a constant learning rate of $5\times10^{-6}$,
weight decay $0.01$, $(\beta_1,\beta_2)=(0.9,0.95)$, and gradient-norm clipping
at $1.0$.

\paragraph{SFT.}
SFT demonstrations contain the assignment acknowledgment, a deterministic
sequence of ascending integers truncated to $k$ tokens, and the correct
response \texttt{Recall: \textless target\textgreater}. We minimize
teacher-forced cross-entropy on assistant tokens only, with the eviction
mask blocking attention from the answer to the assignment turn.
We use batches of $128$,
AdamW with zero weight decay, $(\beta_1,\beta_2)=(0.9,0.95)$, and gradient-norm
clipping at $1.0$. The learning rate warms up for $20$ updates to $10^{-5}$,
then follows a cosine schedule to $10^{-6}$.

\paragraph{Self-distillation (SD).}
SD reuses the SFT demonstrations and adds a KL loss to supervised
cross-entropy, each with coefficient $1.0$. The student uses the eviction mask.
The teacher is a full-context pass of the same current model, without
gradients, and can attend to the assignment. We recompute the teacher
distribution as training progresses. The KL loss approximates
$D_{\mathrm{KL}}(p_{\mathrm{student}}\Vert p_{\mathrm{teacher}})$ over the
student's top $100$ vocabulary tokens at each assistant position, without
renormalizing their probabilities. Optimizer and batch settings match SFT,
except that the learning rate warms up for five updates to $5\times10^{-7}$
and follows a cosine schedule to $10^{-7}$.

All four objectives use the same eviction mask: answer tokens can attend to
the decoded sequence's KVs but not to the assignment turn.

\Needspace{0.5\textheight}
\subsection{Additional Retrieval Results}
\label{app:retrieval-results}
\begin{figure}[H]
  \centering
  \includegraphics[width=\linewidth]{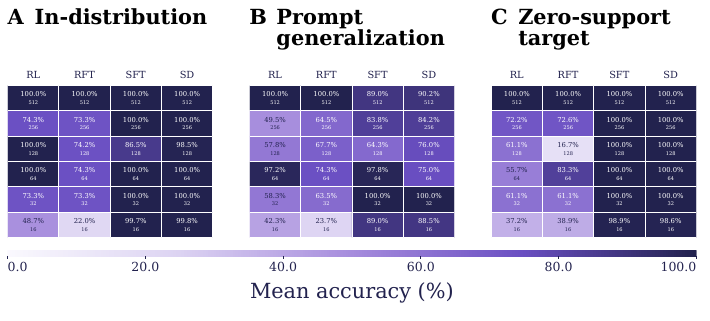}
  \caption{Mean post-eviction accuracy over three seeds: (A) in-distribution,
  (B) prompt generalization, and (C) a zero-support target (TV trained but
  unlisted). Panel C averages all six targets. Smaller cell labels give
  KV-cache token capacity $k$.}
  \label{fig:post-eviction-evaluations}
\end{figure}

\begin{figure}[H]
  \centering
  \includegraphics[width=0.658\linewidth]{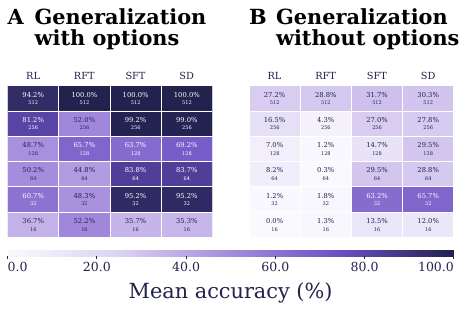}
  \caption{Generalization to new actor names with (A) listed options and
  (B) no listed options. Cells show mean retrieval accuracy over three seeds;
  smaller labels give KV-cache token capacity $k$.}
  \label{fig:post-eviction-generalization}
\end{figure}

Figures~\ref{fig:post-eviction-evaluations} and~\ref{fig:post-eviction-generalization}
show mean accuracy over three seeds. For each condition, we select the learning
rate with the highest mean among configurations with three completed seeds.
All three seeds use that rate. Figure~\ref{fig:post-eviction-512} uses the same
selection rule for RL and SFT. For the TV task, accuracy is averaged over
all six objects, not just TV trials.

\FloatBarrier

\Needspace{0.5\textheight}
\section{Prefill Token Cost}
\label{app:reprefill}

A rollout generates $N$ tokens with compaction operator $\kappa$ and token
budget $B$. Each compaction retains $r = o + s$ tokens: overlap $o$ and any
generated summary $s$. Re-prefill compaction copies them into a new trace and
prefills them again.

\paragraph{Number of compactions.} After compaction, the context holds $r$
tokens, leaving room for $b = B - r$ new tokens. Generating $N$ tokens takes
\[
  m \;=\; \frac{N}{b} \;=\; \frac{N}{B - r}
\]
compactions.

\paragraph{Prefill cost.} Each compaction prefills the $r$ retained tokens,
adding a total token cost of
\[
  \Dre \;=\; m\,r \;=\; \frac{N\,r}{B - r} \;=\; \frac{N(o+s)}{B-(o+s)} .
\]
The cost diverges as $r$ approaches $B$. It depends on the total retained
length, not the split between overlap and summary. Retaining only a summary
of length $s = B/4$ gives $r = B/4$ and $\Dre = N/3$, regardless of $B$.
KV-streams reuses the retained KVs, so $\Dre = 0$.

\end{document}